\documentclass[copyright,creativecommons]{eptcs}
\providecommand{\event}{ICLP 2026} 

\usepackage{iftex}

\ifpdf
  \usepackage{underscore}         
  \usepackage[T1]{fontenc}        
\else
  \usepackage{breakurl}           
\fi

\usepackage{mathptmx}
\usepackage{url}
\usepackage{graphicx}
\usepackage{amsmath,amssymb,amsfonts}
\usepackage{xcolor}
\usepackage{booktabs}
\usepackage{listings}
\usepackage{enumitem}
\usepackage[linesnumbered,ruled,vlined]{algorithm2e}
 \SetAlFnt{\small} 
 \SetAlCapFnt{\small} 
 \SetAlCapNameFnt{\small} 
\usepackage{subcaption}
\usepackage{multirow}

\definecolor{myblue}{HTML}{1F77B4}

\SetCommentSty{mycommfont}

\usepackage{listings}
\title{Explaining Weather Bulletins via ILP~\thanks{Research partially supported by Interdepartment Project on AI (Strategic Plan UniUD–22-25). A.Dal~Pal{\`{u}}, A.Dovier, T.Dreossi, A.Formisano, and E.Santi are members of GNCS-INdAM, Gruppo Nazionale per il Calcolo Scientifico. E.Santi is supported by FSE/FVG PhD Grant  on 
``XAI-FVG Explainability of Weather
Forecasting in FVG'' 
(G23C23001130008).}
}
\author{\quad Enrico Santi \qquad\qquad\qquad\qquad\quad
Alessandro Dal Pal\`{u}
\institute{University of Udine, DMIF, Udine, Italy\qquad\qquad\quad University of Parma, SMFI, Parma, Italy} \email{\quad enrico.santi@uniud.it  \qquad\qquad\qquad   alessandro.dalpalu@unipr.it}
\and
 Agostino Dovier \qquad
 Talissa Dreossi \qquad
 Andrea Formisano
\institute{University of Udine, DMIF, Udine, Italy} \email{name.surname@uniud.it} 
}

\newcommand{\titlerunning}{Explaining Weather Bulletins via ILP}
\newcommand{\authorrunning}{Santi, Dal Pal{\`{u}}, Dovier, Dreossi, Formisano}

\newtheorem{example}{Example}[section]
\newtheorem{definition}{Definition}[section]

\begin{document}
\maketitle

\begin{abstract}
Inductive Logic Programming (ILP) originated within the Logic Programming community in the Nineties as a framework for combining symbolic learning with declarative knowledge representation. Nowadays, mature ILP frameworks exist and they are capable of learning complex, non-monotonic hypotheses, thus broadening both the modeling capabilities and the scope of real-world applications of ILP.
This work is primarily based on the FastLAS2 framework and aims to generate simple, interpretable hypotheses to help clarify the weather bulletins issued by OSMER FVG, the Regional Meteorological Observatory of the Italian region of Friuli Venezia-Giulia. 
In this paper we present a pipeline that, starting from simulated meteorological raw data and from OSMERs' bulletins (used as ground truth), extracts data as ASP facts and generates ILP examples. From such examples an explanatory hypothesis is then inferred via FastLAS2. 
Such a hypothesis (translated into natural language) explains the weather forecast issued by human experts, and in particular the rationale behind experts' choices of specific symbols in the bulletin pictogram (the symbol-annotated meteorological map of the forecast).
The proposed approach is general, not specific to any particular region and 
it can equally be applied to bulletins from other sources and to different regions.
\end{abstract}

\begin{figure}[t]
	{    \centering
    \begin{minipage}{.55\textwidth}
    \includegraphics[width=\linewidth]{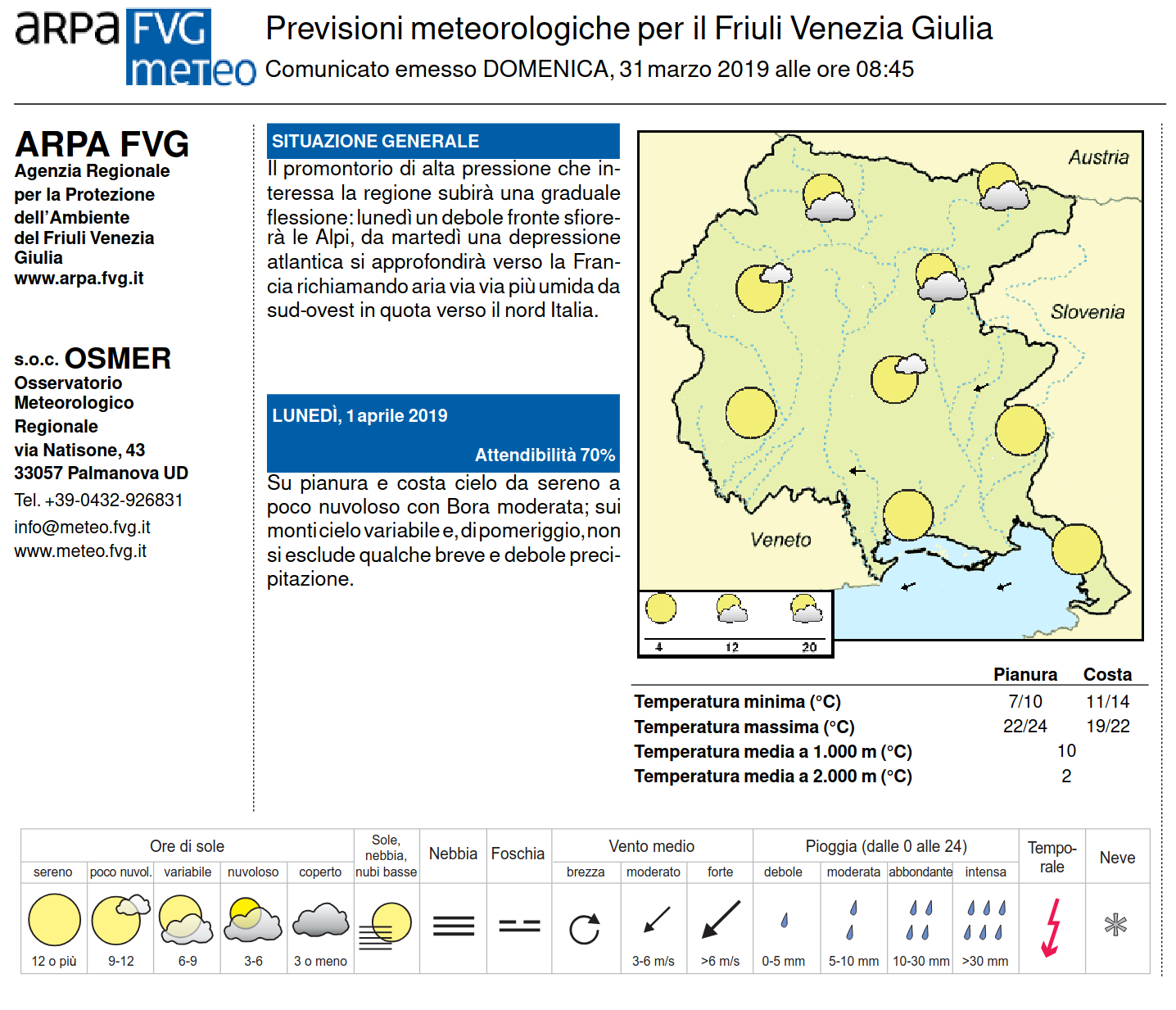}
    \end{minipage}
    \begin{minipage}{.3\textwidth}{\scriptsize
{\bf General Info}\\
The high-pressure ridge influencing the region will gradually weaken. On Monday, a weak frontal system will affect the Alpine area. From Tuesday onward, an Atlantic low will move toward France, advecting increasingly humid southwesterly air at upper levels over northern Italy.
\\
{\bf Monday, April 1, 2019}\\
Clear to partly cloudy skies, with moderate Bora winds over the plains and along the coast.
In the mountainous areas, variable cloudiness; brief and light precipitation cannot be ruled out in the afternoon.
  }  \end{minipage}
    \caption{\label{fig:bullettin}The bulletin for the 1st of April 2019 for the FVG region issued by OSMER. It includes a pictogram describing the sky coverage and precipitation in different key locations. 
The report also includes a brief textual description summing up some of the reasons the domain experts considered for the forecast.
English translation is reported in the right box.}
	}
\end{figure}

\section{Introduction}

Weather forecasting is essential for anticipating and reducing the impact of severe weather events. It plays a key role in several fields, from agriculture to transportation. People use forecast reports and bulletins for planning their routines, vacations, and taking informed decisions. 

Forecast bulletins are required to be concise and clear. They typically present a well-defined structure that arranges graphical elements, tables and short explanatory text related to the forecast of a geographic region. As an example, Figure~\ref{fig:bullettin} shows the structure of an OSMER (OSservatorio MEteorologico Regionale del Friuli Venezia-Giulia) weather report for a single day. 

Traditional approaches to weather forecasting, such as numerical weather prediction (NWP) models~\cite{NWP}, are based on mathematical formulations that simulate atmospheric fluid dynamics, which, unfortunately, can be computationally expensive. To mitigate this limitation, NWP models~\cite{Kalnay2002} are increasingly complemented by statistical and machine learning techniques~\cite{rasp2018deep,mcgovern2017using,Liu2024,zhou2019forecasting,zhang2024weather}.

\begin{figure}[t]
	{\centering
{\includegraphics[width=0.7\linewidth]{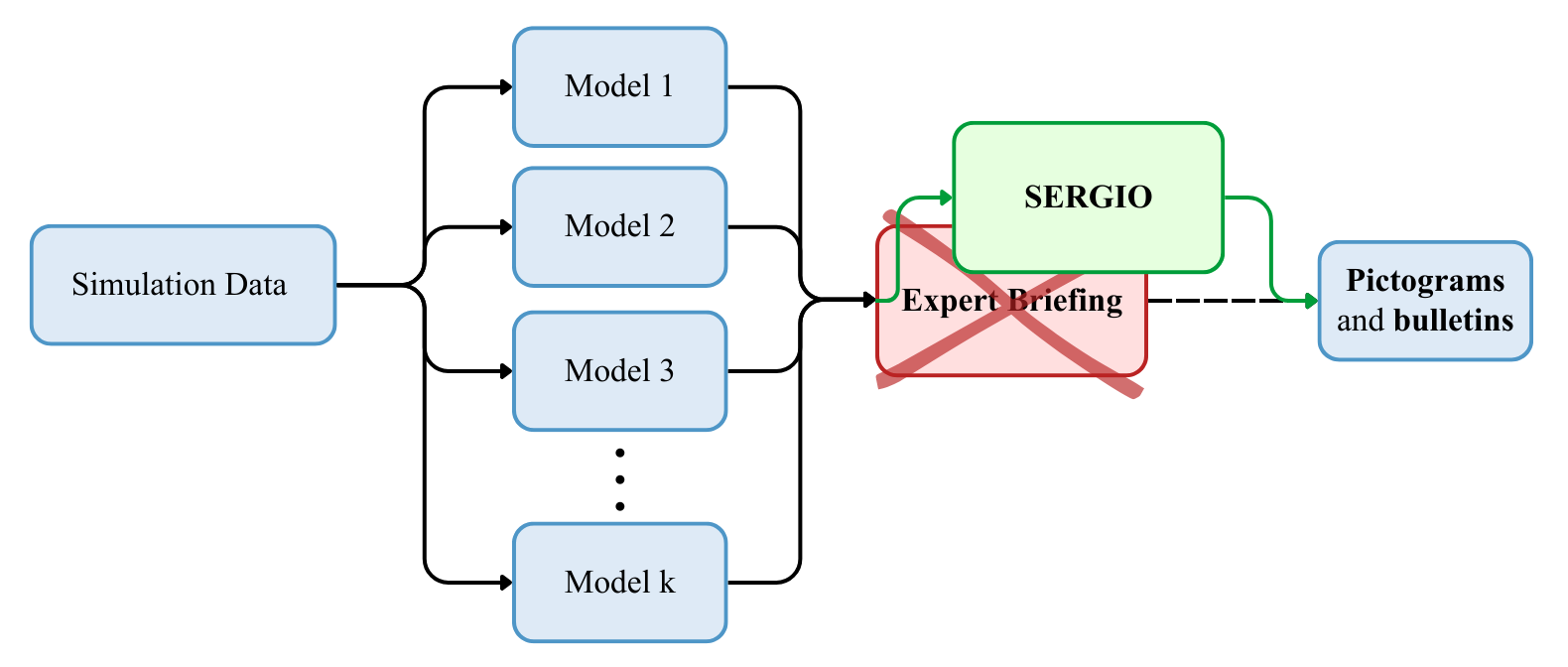}
    \caption{The overall aim of the proposed work. The current OSMER weather forecasting process includes several Deep Learning models that are used for prediction (as black boxes). After the predictions, a group of experts (physicist experts in weather forecast and experts in communication) decides how to label the map and creates the sentences to be written in the bulletin. Our tool SERGIO ({\bf S}imulating {\bf E}xplanation of {\bf R}e{\bf GIO}nal weather forecast) could  replace this final stage.\label{fig:sergio}}
	}}
\end{figure}

OSMER's workflow for issuing weather forecasts and bulletins also benefits from the integration of deep learning techniques.  The process, summarized in Figure~\ref{fig:sergio}, involves a number of deep learning models that are used as ``black boxes'' to generate a set of (not necessarily agreeing) predictions about the weather in a specific geographical zone.  A group of communications experts and physicists with expertise in weather forecasting examines and evaluates such predictions to synthesize them into a single forecast.  
They then decide on the labels for the map areas and draft the text that will accompany the bulletin.

Exploiting deep learning techniques, such as neural networks, in weather forecasting achieves high predictive accuracy but lacks explainability and interpretability. This limits user trust and makes domain experts a crucial component for validating and refining models' outputs when producing the final weather reports \cite{talissaWeateher,murakami2021generating,BELZ2008,sukhorukov2025}.

This work explores FastLAS2 \cite{fastlas}, an Inductive Logic Programming framework \cite{Muggleton91}, as a mean to generate rules that explain weather bulletins. FastLAS2 is a fast and relatively scalable framework that has been proven to be able handle some large datasets and complex background knowledge, using a scoring function to identify optimal solutions \cite{drozdov2021online,law2022search}. It has also been used with the help of neural networks \cite{dreossi2025explainable,charalambous2023neuralfastlas}. In particular, in our previous work \cite{talissaWeateher}, we investigated the use of FastLAS, for explainable weather forecasting. The proposed approach leverages FastLAS to learn rules for predicting rain levels. An empirical evaluation conducted on real data (from OSMER) demonstrated that FastLAS is capable of learning models that explain rainfalls from relatively small datasets, achieving performances similar to classical machine learning baselines (SVM, decision tree, random forest).
Note that the previous work \cite{talissaWeateher} employed FastLAS for a prediction task, as the rain data were not generated by any black-box model. In contrast, the use of FastLAS in the proposed approach is aimed at an explanatory task, i.e., not at predicting what has happened, but rather at learning rules that explain why black-box models produced such predictions. 
Moreover, \cite{talissaWeateher} considers single, unrelated cities, whereas the approach described in this paper integrates data from multiple nearby locations. This is motivated by the fact that weather conditions in surrounding areas may help explain the weather observed at a given location.

Besides FastLAS, another ILP framework, FOLD-RM~\cite{foldrm}, was initially considered for use with the data extracted from the pipeline. However, we decided not to report the details because, although FOLD-RM is very intuitive to use, it lacks support for user-defined cost functions to guide the hypothesis search. More importantly for our scenario, it does not allow the use of rules as background knowledge.

\smallskip

The paper is organized as follows:
Section \ref{sec:ILP} recalls some background on ILP, focusing specifically on the FastLAS2 framework \cite{fastlas}. Section \ref{sec:pipeline} describes the implementation of the pipeline which extracts ASP facts from raw simulations on meteorological data and the bulletin. Section \ref{sec:model} presents the ILP model implemented in FastLAS2 in order to derive for a pictogram our hypothesis, then the experiments we ran to train and validate our model are discussed. Lastly, in Section \ref{sec:conclusion} we draw our conclusions.  Source code, test instances, and results are available from \url{https://clp.dimi.uniud.it/sw}.

\section{Inductive Logic Programming} 
\label{sec:ILP}
The main goal of the machine learning branch of Inductive Logic Programming is to learn logical rules from \emph{examples} and \emph{background knowledge}. The process of learning consist of searching an \emph{hypothesis space} for rules that are coherent with prior knowledge and that best explain the instances (examples) provided.
The work presented in this paper is based on the \emph{Learning from Answer Sets} (LAS) ILP framework called FastLAS2 \cite{fastlas}, as it uses ASP which is especially useful for learning non-monotonic logical rules.  In LAS, a learning problem is called a task, which is formalized by the tuple $T = \langle B, S_M, E \rangle$, where $B$ denotes the background knowledge, namely a set of ASP rules, $S_M$ is the hypothesis space, i.e.\ the set of rules that can possibly be learned, and $E$ which is the set of examples that must be explained by the knowledge base extended with learned rules.
The hypothesis space can be either explicitly provided as input (that is, a set of hypotheses formalized as rules to be filtered/checked) or implicitly defined using \emph{mode bias}.
A mode bias is a pair $\langle M_h, M_b \rangle$, where $M_h$ and $M_b$ are sets of so-called ``mode declarations'' for the heads and bodies of rules. 
$M_h$ defines the name of the predicates that can be used as head, the number of arguments, and the scope of the
arguments.
$M_b$ lists the names and the arguments of the predicates that \emph{can} be used as body atoms.
If $M_b$ contains $h$ predicates and $M_h$ involves a single predicate, then they describe an exponential number ($2^h$) of rules in the hypothesis space~$S_M$.
The LAS solver is also permitted to explore the insertion, in the body of the learned rules, of comparison predicates (e.g., $X < 5$ or $Y\: {!}{=}\: Z$) among variables occurring in the body.

\begin{example}
Suppose we would like to infer the predicate \lstinline{forecasted_sky/2} with
the first argument defined by a predicate 
\lstinline{location/1} (for example, by the facts \lstinline{location(gorizia)},
\lstinline{location(udine)}, \lstinline{...}) and having as second argument 
a numerical value bounded by
the predicate \lstinline{coverage/1} (as in, for instance, \lstinline{coverage(0..6)}).
Thus, we express $M_h$ as:
\begin{lstlisting}
#modeh(forecasted_sky(const(location),const(coverage))).
\end{lstlisting}

\noindent
Suppose, moreover,  we conjecture that the predicate  \lstinline{forecasted_sky/2} might depend on the 
predicate \lstinline{city_covered_less_than/2} (that 
can be extracted from the data generated by the numerical weather simulations by counting the number of hours a cloud covers a location) and on the predicate
\lstinline{wind_blowing/3} that has to do with the direction and speed of the
wind at a given location (direction and speed have bounds defined by the predicates
\lstinline{const(w_dir)} and \lstinline{const(w_speed)}). This can be expressed by including the following modes in~$M_b$:
\begin{lstlisting}
#modeb(city_covered_less_than(const(location),const(hours)))
#modeb(wind_blowing(const(location),const(w_dir),const(w_speed)))
\end{lstlisting}
\end{example}

Examples are \emph{partial interpretations}: a partial interpretation $e_{pi}$ is a pair of sets of ground atoms $e_{pi}=\langle e^{inc}, e^{exc} \rangle$, where $e^{inc}$ represents the inclusions, that is, atoms that must be true, and $e^{exc}$ represents the exclusions, or atoms that must be false when inclusions are true. When a candidate interpretation $I$ satisfies the conditions  $e^{inc} \subseteq I$ and $e^{exc} \cap I = \emptyset$, it is said to \emph{extend} the partial interpretation $e_{pi}$. 

In FastLAS2 examples can be also context-dependent.
Precisely, a \emph{context-dependent partial interpretation} (CDPI) is a tuple $e_{cdpi} = \langle e_{pi}, e_{ctx} \rangle$, where $e_{ctx}$ is an ASP program that represents the context in which the example should be considered and $e_{pi}$ is a partial interpretation. If there is at least one stable model $A$ of $P \cup e_{ctx}$ such that $A$ extends $e_{pi}$, then a program $P$ is said to accept the CDPI $e_{cdpi}$. 

FastLAS2 also permits to define a \emph{score function} which is used, along with example's penalty, to calculate the total cost of a solution. In particular, the score function is user-defined by specifying penalties for predicates in the head or in the body of the learned rules.
Real-world data is inherently \emph{noisy}, in the LAS framework, noise can be modeled by associating examples with weights, referred to as \emph{penalties}. A \emph{weighted context-dependent partial interpretation} (WCDPI) $e$ is defined as a tuple $\langle e_{id}, e_{pen}, e_{cdpi} \rangle$, where $e_{id}$ is an identifier for $e$, $e_{pen}$ is a positive integer denoting the \emph{penalty}, and $e_{cdpi}$ is a context-dependent partial interpretation. A LAS task is said to be \emph{noisy} when its examples are WCDPIs.
\begin{definition}[Inductive solution]\label{def:opl}
  A \emph{noisy LAS task} is a tuple $T^{noise}=\langle B, S_M, E\rangle$ where $B$ is an ASP program, $S_M$ is the hypothesis space, and $E$
  is a finite set of WCDPIs. A hypothesis $H \subseteq S_M$ is an inductive
  solution of $T^{noise}$ iff $\forall e \in E$,
  $B\cup H$ accepts $e$. 
  Let $\mathcal{T}^{noise}$ denote the set of all noisy LAS tasks.
\end{definition}

If an hypothesis does not accept an example, it \emph{pays the penalty} associated with that example. Informally, penalties are used to compute the \emph{cost} incurred by a hypothesis for failing to explain given examples. The overall cost function of a hypothesis is defined as the sum of the penalties of all examples not accepted by the hypothesis, augmented by a user defined scoring function. A \emph{scoring function} is a function $\mathcal{S} : \mbox{ASP} \times \mathcal{T}^{noise} \rightarrow \mathbb{R}_{>0}$, where ASP denotes the set of all possible ASP programs which can be generated by rules in $S_M$. The objective of a noisy LAS task is to identify a hypothesis that minimizes the overall cost function over a given hypothesis space with respect to a set of WCDPI examples. Such a hypothesis is referred to as \emph{optimal}.
FastLAS2 is a noisy LAS system that supports user-defined scoring functions, enabling domain-specific optimization criteria to guide the search process.

\section{From Meteorological Data to ASP Facts}
\label{sec:pipeline}
This section describes the automated pipeline that processes raw data (simulation data and pictograms from bulletins provided by experts). We describe how to identify and convert to ASP facts relevant features that are at the basis of experts' reasoning for crafting a weather prediction. The facts obtained will constitute the knowledge base provided to FastLAS2 for the learning phase.

As recalled in Section \ref{sec:ILP}, an example, i.e.\ a WCDPI, is composed of the partial interpretation and a context.
In our application, such facts can be derived from different data sources. 
For each day of analysis, the context facts ($e_{ctx}$) are computed by abstracting some of the data from the Copernicus European Regional ReAnalysis (CERRA) dataset, which contains simulated data that match the meteorological evolution of physical measurements for the specific day. The atmospheric variables considered, as well as generated facts, are described in Section~\ref{sub:cerra}.
Another source of data is the daily OSMER bulletins. We extrapolate pictograms and produce high level facts that feed the partial interpretation $e_{pi}=\langle e^{inc}, e^{exc}\rangle$. 

The dataset used in our work, as a proof of concept of a general methodology, focuses on the region Friuli Venezia-Giulia (FVG), in the north-east of Italy.

\begin{figure}[tb]
    \centering
    \includegraphics[width=0.85\linewidth]{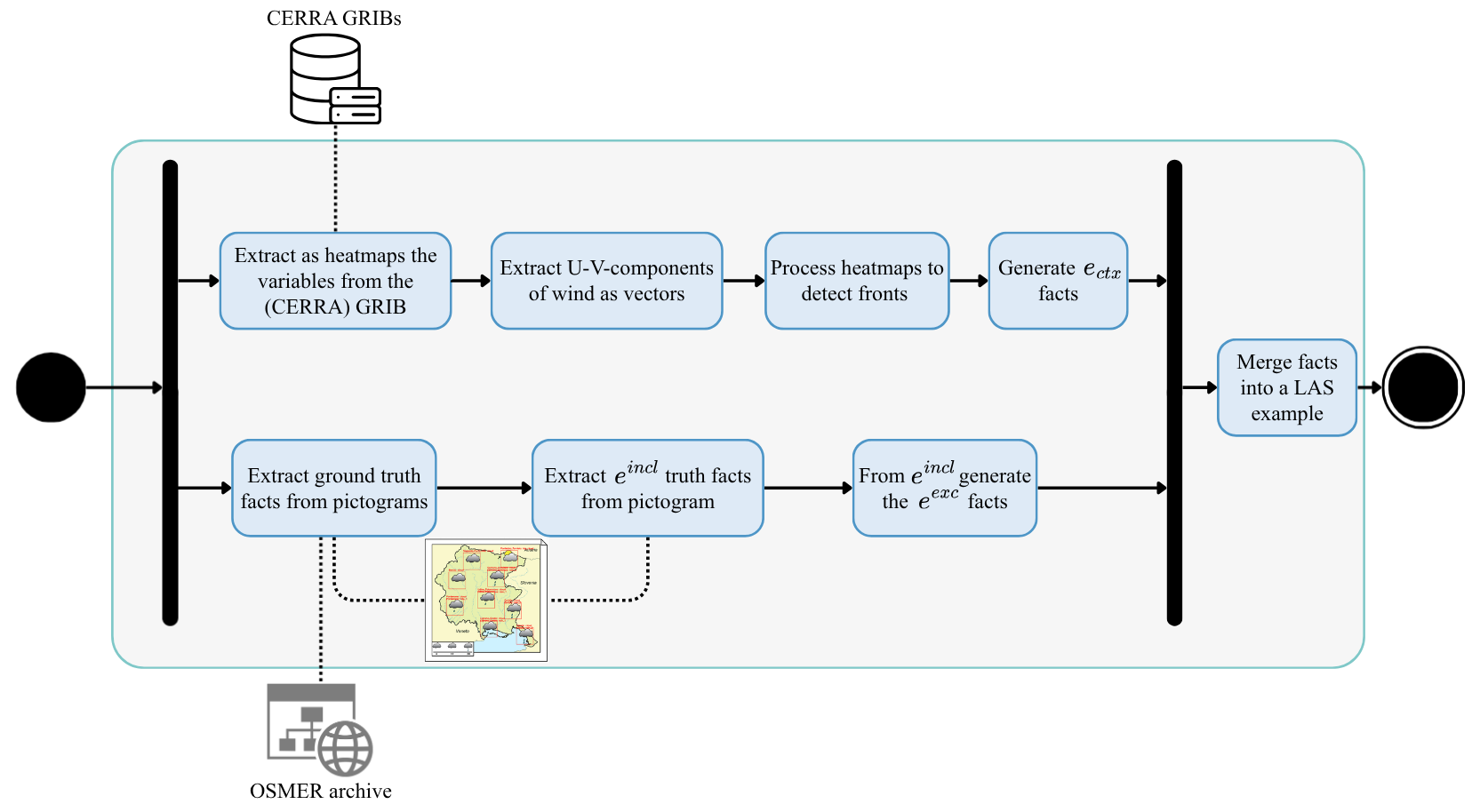}
    \caption{An activity diagram describing the process used to generate each WCDPI example. Note that for the WCDPIs the context facts ($e_{ctx}$) and the partial interpretations $e_{pi}$ can be computed independently since they are derived from different data sources.}
    \label{fig:activity}
\end{figure}

The overall pipeline is shown in Figure~\ref{fig:activity} and described in what follows.

\subsection{Ground Truth, Pictogram Facts Extraction}
\label{sub:pictogram}
In order to generate the WCDPI for each daily example, we process bulletins provided by OSMER\footnote{Available at:\url{https://www.osmer.fvg.it/archivio.php}.}.
In detail, we extract the pictogram describing the overall situation for the day in the FVG region (see Figure~\ref{fig:bullettin}). We identify, through image analysis, nine icons located across the maps and extract the qualitative sky coverage (e.g., \emph{clear}, \emph{partly~cloudy}, \emph{variable}, \ldots) and precipitations (from 0 to 6 water droplets) in the area.

\begin{figure}[tb]
    \centering
    \includegraphics[width=0.75\linewidth]{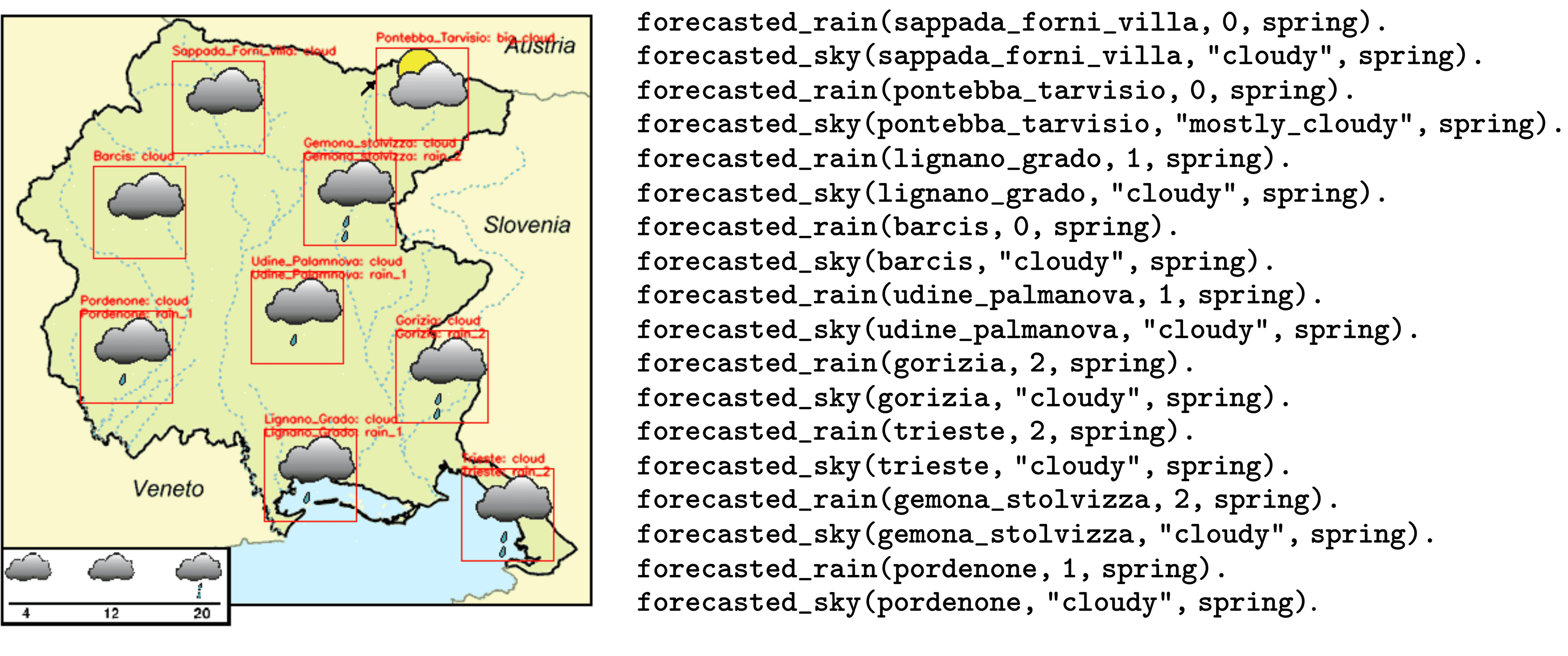}
    \caption{On the left the result of the multi-scale template matching on an OSMER pictogram. On the right the extracted facts for the different locations. The facts included describe sky coverage and precipitation levels at the locations depicted by the pictogram, also including the season. }
    \label{fig:einc}
\end{figure}

Given a date, in order to derive $e^{inc}$, a multi-scale template matching algorithm on the pictogram is applied. The templates include all possible icons which can be used in the pictograms.
For each of the nine locations, two predicates are generated, \lstinline{forecasted_sky/3} which describes the predicted cloud coverage and \lstinline{forecasted_rain/3} describing the predicted rain level. The rain level ranges from 0 (no rain at the location) to 6 according to the number of droplets in the icon at the location. The coverage argument of the forecasted sky predicate can be one in 
$\{$\lstinline{cloudy}, \lstinline{partly_cloudy}, \lstinline{mostly_cloudy},
\lstinline{mostly_clear}, \lstinline{sunny}$\}$
according to the type of icon at the location.
The last parameter of both predicates is the season (the reason for this choice is discussed in Section \ref{sec:model}).
Figure~\ref{fig:einc} presents an example of the analysis: on the left the result of the multi-scale template matching on a pictogram and on the right the derived ASP facts. 


\subsection{CERRA Facts Extraction}
\label{sub:cerra}

CERRA (CERRA sub-daily regional reanalysis data for Europe on pressure levels\footnote{Available at: \url{https://cds.climate.copernicus.eu/datasets/reanalysis-cerra-pressure-levels}.}) is a high-resolution regional reanalysis dataset providing reconstruction of different atmospheric and surface meteorological variables over Europe.
While sharing the same format of weather simulation models, this dataset aims at reproducing \emph{a-posteriori} the most likely data that evolved in the observed events. It is considered to be more reliable than simulations and it represents a convenient test bench for our purposes.
This dataset is encoded with the format GRIB (General Regularly-distributed Information in Binary form), a standardized operational meteorological data format, designed to store grid data generated by numerical weather forecast models \cite{grib}.
Each GRIB file contains one or more independent records, each consisting of metadata and binary data values for a specific variable on a defined grid and time step.

The CERRA dataset contains several atmospheric variables such as \emph{cloud cover}, \emph{relative humidity} and \emph{temperature} at different heights, from 1000hPa to 1hPa.
Such data is provided at multiple spatial and temporal resolutions. In this work, we use the high-resolution reanalysis, which is available at a horizontal resolution of 5.5 km and a temporal resolution of 1 hour.
To derive the context of each example we considered the following variables, at six different height levels: 
$$\begin{array}{rl rl} 
1000 hPa & (\simeq 100/110m) & 925 hPa  & (\simeq 750/900m)\\
850 hPa & (\simeq 1500m) & 700 hPa & (\simeq 3000m)\\ 
500 hPa & (\simeq 5500m) & 300 hPa & (\simeq 9000m)
\end{array}$$
\begin{itemize}
    \item \emph{Cloud coverage}: Percentage value of cloud coverage within a grid box.
    \item \emph{Relative humidity}: Value describing the water vapor pressure as a percentage of the value at which the air becomes saturated.
    \item \emph{Temperature}: The air temperature in Kelvin.
    \item \emph{U-component of wind}: The East-West component of the horizontal wind velocity.
    \item \emph{V-component of wind}: The North-South component of the horizontal wind velocity.
\end{itemize}
All the variables considered were extracted over the area bounded by $10$ and $15$ degrees of East longitude and $44.5$ and $48$ degrees of North latitude. 
Such a choice was driven by the need to capture the whole FVG region, together with a surrounding context area (a portion of the Veneto region at East, a portion of Austria, including Salzburg and Graz, at North, and part of Slovenia at West).

We process raw data and retrieve planar rectified images for each scalar variable and two maps for wind, represented by a 2D vectorial information (U and V components). Data is sampled at each hour of the day, therefore we collect 24 heatmaps for each of the 6 heights.

Following common practices, we analyze those images and identify blob-like regions that differentiate from the adjacent regions. Such features are an abstraction of coherent configurations of meteorological variables that can be exploited in higher level reasoning. We focus on clouds concentrations, high humidity and temperature blobs.
As further abstraction, after detection of blobs we also track their movements across time at the same altitude, by means of the framework TOBAC 1.2 that is capable of detecting and tracking individual clouds from different image sources, such as images from cloud-resolving model simulations and geostationary satellite retrievals \cite{tobac}.
In order to cope with noise in the images, which causes TOBAC to deal with hundreds of noisy blobs, we quantize the values of each image to a few discrete levels (e.g.,~5). The smoothed version of the new image has fewer features, while preserving the ones that range over larger spatial scale.

\begin{figure}[tb]
    \centering
    \includegraphics[width=0.9\linewidth]{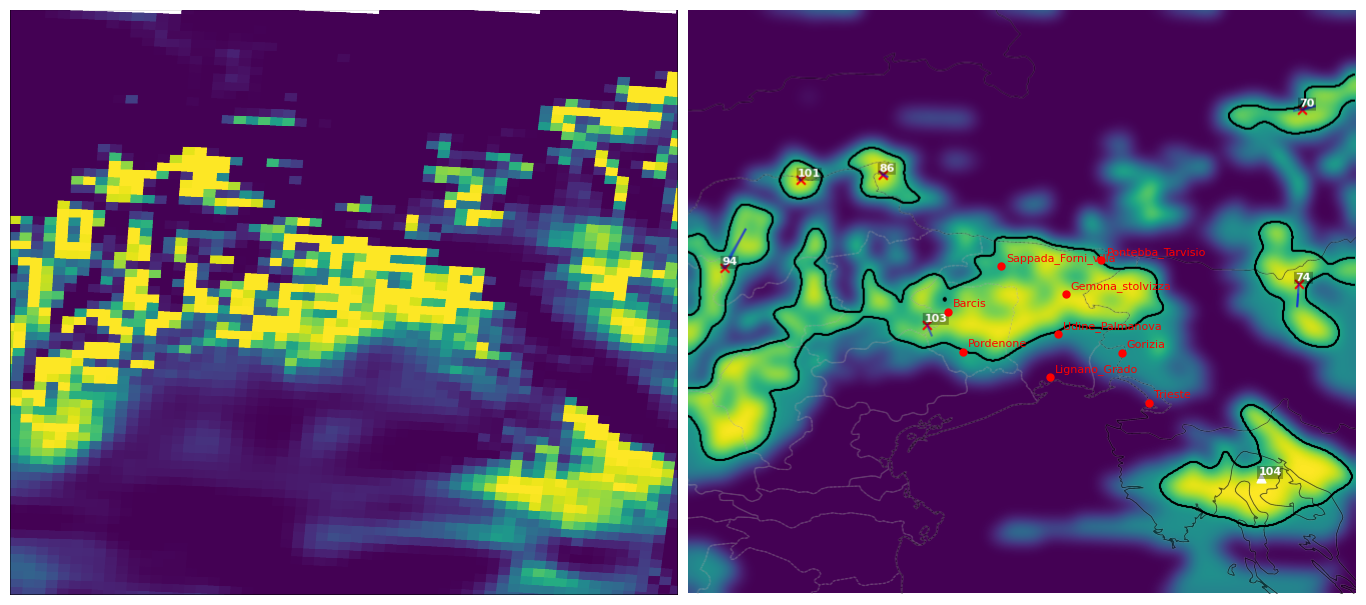}
    \caption{On the left a heatmap, related to cloud coverage data extracted from a GRIB file. On the right the same heatmap where a three level quantization and a Gaussian filter has been applied. The regions (clouds) detected by TOBAC are outlined in black, each cloud presents an id. For geographical reference location names and political borders have been added on top.  }
    \label{fig:cloudexample}
\end{figure}

\begin{figure}[tb]
    \centering
    \includegraphics[width=0.9\linewidth]{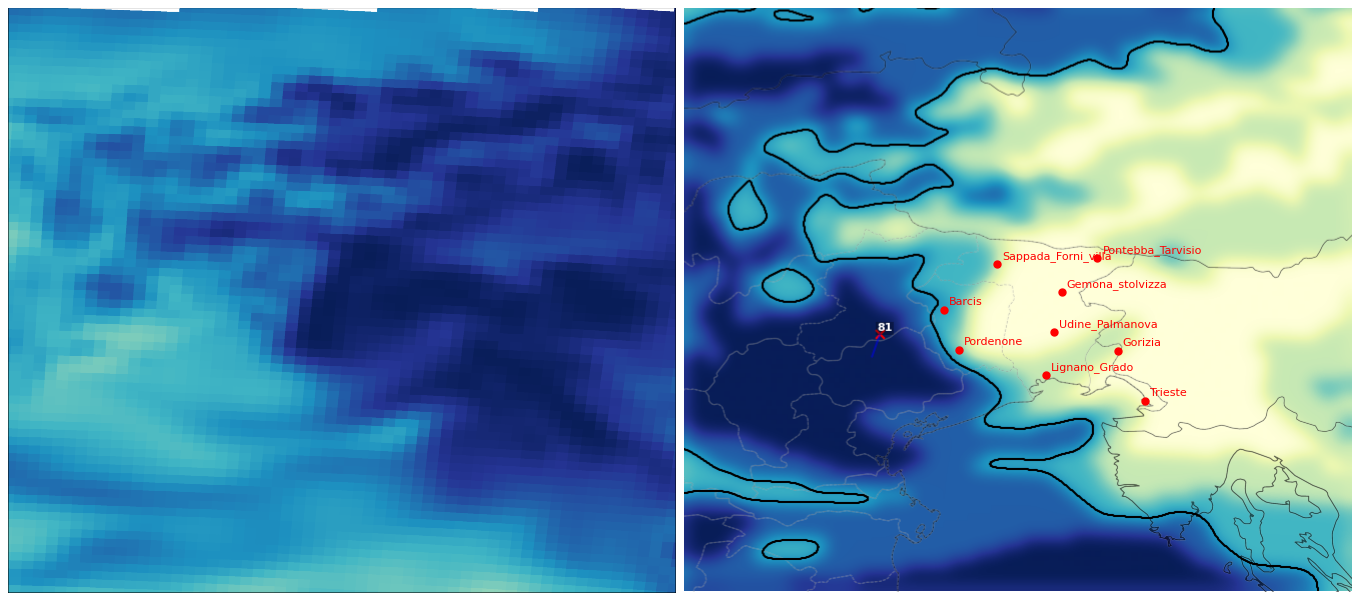}
       \caption{On the left a heatmap, related to humidity data extracted from a GRIB file. On the
    right the same heatmap where a five level quantization and a gaussian filter has been applied. The high humidity front detected by TOBAC is outlined in black, note that smaller blobs are not detected due to the higher threshold used for humidity and temperature. For geographical reference location names and political borders have been added on top. }
    \label{fig:front}
\end{figure}

Domain experts suggested that fronts (large and elongated blobs) are more relevant than small blobs (even if more intense), when considering humidity and temperature heatmaps. 
Therefore, for those images we quantized 5 levels but filtered out small blobs. For cloud coverage, we instead quantized only 3 levels but retained also smaller blobs.
Figure~\ref{fig:cloudexample} (resp., \ref{fig:front}) presents an example of a cloud (resp., humidity) heatmap extracted from a GRIB file and the corresponding image after quantization and Gaussian filtering, to which TOBAC was applied.
GRIB data for wind is sampled at very coarse resolution. Therefore we interpolate the data to match the same resolution associated to the other maps, we then generate the images containing directions and magnitudes of the wind vectors.

Facts extraction is based on the analysis of images (classified by time and height) and the output of TOBAC tracking. We collect the coordinates of the nine target locations in the images to extract punctual details. Moreover, front analysis is performed. We describe how each type of facts is retrieved:
\begin{itemize}
	\item 
    Facts \lstinline{cloud_at_H_covers/2} (where \lstinline{H} is a height level) describe
    cloud coverage at each location: the first argument is one of the nine locations while the second is the hour (\lstinline{0..23}) at which the cloud (at level \lstinline{H}) has been detected over the location.
    \item  
Two facts \lstinline{wind_blowing_morning/3} and \lstinline{wind_blowing_afternoon/3} summarize wind data at each location.
    For both predicates the arguments are a location, a direction (\lstinline{"N"}, \lstinline{"NW"}, \lstinline{"W"},\lstinline{"SW"}, \lstinline{"S"}, \lstinline{"SE"}, \lstinline{"E"}, or \lstinline{"NE"}) and integer speed (km/h). The speed (resp., direction) is an average (resp., weighted average) of the values extracted from GRIB data at different heights during the morning or afternoon hours.
    \item For temperature and humidity we model additional predicates.
    First, for each location the predicates \lstinline{temperature_at_morning/2}, \lstinline{temperature_at_afternoon/2},     
    \lstinline{humidity_at_morning/2}, and \lstinline{humidity_at_afternoon/2} are generated. The arguments of such predicates are a location name and an integer value. For the temperature the value is the average temperature over a location, while for the humidity it describes the percentage of humidity. Such values are extracted from raw GRIB data, to avoid loss of information.  
    Additional facts describe the daily trend of temperature and humidity:
\lstinline{temperature_increased_at_afternoon/1},  \lstinline{temperature_decrease_at_afternoon/1}, 
    \lstinline{humidity_increased_at_afternoon/1},  \lstinline{humidity_decreased_at_afternoon/1}.
The argument of these last facts is a location.
\item For both temperature and humidity we  generate facts about fronts:
        \lstinline{hum_front_morning_at_H/2}, \lstinline{hum_front_afternoon_at_H/2}, 
        \lstinline{temp_front_morning_at_H/2}, and \lstinline{temp_front_afternoon_at_H/2} (where \lstinline{H} is a height level). They are generated when TOBAC detects, for more than two consecutive hours, a humidity or temperature difference sufficiently widespread to be considered a front. Both parameters are locations: we keep track of the same front that traverse pairs of locations cities during morning/afternoon. 
\end{itemize}

The following is an example of a generated $e_{ctx}$, that is, the set of facts that make up the context of a day. Note that all data present in each context is automatically derived from CERRA GRIBs. 

\begin{lstlisting}[basicstyle=\ttfamily\upshape\footnotesize]
% cloud coverage:
cloud_at_100m_covers(pontebba_tarvisio, 1).
cloud_at_100m_covers(pontebba_tarvisio, 2).
cloud_at_750m_covers(pontebba_tarvisio, 1).
...
cloud_at_1_4km_covers(pontebba_tarvisio, 0).
cloud_at_1_4km_covers(pontebba_tarvisio, 1).
cloud_at_1_4km_covers(pontebba_tarvisio, 2).
...
cloud_at_3km_covers(pontebba_tarvisio, 5).
cloud_at_3km_covers(gemona_stolvizza, 6).
...
cloud_at_9km_covers(trieste, 2).

% summing up temperature and humidity facts (one per location):
temperature_at_morning(sappada_forni_villa, 278).
temperature_at_afternoon(sappada_forni_villa, 279).
temperature_at_morning(pontebba_tarvisio, 279).
temperature_at_afternoon(pontebba_tarvisio, 277).
...
temperature_at_morning(pordenone, 277).
temperature_at_afternoon(pordenone, 276).
humidity_at_morning(sappada_forni_villa, 34).
humidity_at_afternoon(sappada_forni_villa, 27).
...
humidity_at_morning(pordenone, 47).
humidity_at_afternoon(pordenone, 66).

% summing up winds and fronts:
wind_blowing_morning(sappada_forni_villa, "SW", 4).
wind_blowing_afternoon(sappada_forni_villa, "S", 4).
...
wind_blowing_afternoon(pordenone, "S", 3). 
temp_front_afternoon_at_750m(gemona_stolvizza, pontebba_tarvisio).
temp_front_afternoon_at_750m(gorizia, pontebba_tarvisio).
temp_front_morning_at_3km(pontebba_tarvisio, sappada_forni_villa).
...
hum_front_morning_at_100m(sappada_forni_villa, udine_palmanova).
hum_front_morning_at_100m(lignano_grado, pordenone).
\end{lstlisting}


\section{ILP Model and Results}
\label{sec:model}
We now present the background knowledge crafted to be used by FastLAS2 to learn the hypothesis. Some implementation choices will be addressed and the obtained results will be discussed.

\paragraph*{Background Knowledge}\label{sub:bg}
The background knowledge is a crucial part in ILP since it allows to intensionally define (through the mode biases---see Section~\ref{sec:ILP}) the search space of the hypotheses and, possibly, to define auxiliary predicates which infer additional facts from each example.

By modifying the background knowledge several variations of the model (i.e., different final hypotheses) have been considered, many of which, during the learning of the hypothesis itself, lead to a learning process exceeding 128GB of RAM, and, consequently, killed. 
This serious memory issue forced us to aggregate data over the different height levels (e.g., the facts related to humidity and temperature---see Section~\ref{sub:cerra}). 
For the same reason, in preparing 
the partial interpretation $(e^{inc},e^{exc})$ of each example, we used single locations and not the set of 9 locations, even if we allowed to use also facts concerning the whole region.
Namely, even though each $e_{pi}$ is related to a single location, the context of each example, $e_{ctx}$, carries all the facts extracted over the whole region for that day.
This choice has been made so that FastLAS, when learning the rules for a specific location, could also leverage facts concerning neighboring locations, if relevant.

Let us focus on the auxiliary predicates that arose after meetings with domain experts that,
in order to predict the sky coverage of a given location, keep into account the sunlight hours, which vary from season to season.
Thus, we included in the background knowledge a predicate to define the range of sunlight hours for each season as well as a predicate deriving the season from a given date (present in the context of each example). 
Such an information, which also shows up in the predicates in $e^{inc}$, allows us to learn more precise hypotheses since for a given location, an explanation for the sky can keep into account alongside the number of hours covered by a cloud also the total number of sunlight hours. For this reason the predicates \lstinline{forecasted_sky/3} and \lstinline{forecasted_rain/3} present a season argument.
$M_b$ contains the predicates listed in Section~\ref{sub:cerra}. Consider a LAS example $e$, the inclusion $e^{inc}$ will describe the forecast for a specific location $\ell$. The predicates in $M_b$ distinguish (with different weights) whether they are used on the location $\ell$ on a neighboring location. Such a choice was made to enable different costs for the predicates, making FastLAS preferring \emph{local} explanations before considering the neighboring locations. 
The following example presents the FastLAS commands to specify a cost of 1 for including in a body of a rule the \lstinline{temperature_increased_at_afternoon} on the location presented in the head of the same rule and a cost of 2 for including the predicate on a neighboring location:
\begin{lstlisting}
#bias( "penalty(2,tmp_inc_n(X)) :- in_body(temperature_increased_at_afternoon_neighbour(X))." ).
#bias( "penalty(1,tmp_inc(X)) :- in_body(temperature_increased_at_afternoon(X))." ).
\end{lstlisting}

\paragraph*{Training}
The training (and evaluation) dataset comprises 70 days of data selected in consultation with the domain experts from OSMER. The selected days span 2025 and cover all four seasons, specifically incorporating relevant seasonal phenomena (e.g., \emph{scirocco winds} and \emph{spring cold fronts}). This expert-guided selection ensures that the learned hypothesis accurately accounts for such phenomena.  We limited the dataset to 70 days for two primary reasons: First, the FVG region is relatively small and, therefore, a period of 70 days provides sufficient data to capture common seasonal phenomena in the region.  Second, even within a small geographical area, 70 days of GRIB data require hundreds of gigabytes of storage. This volume strains the capacity of our prototype pipeline, which must process the entire dataset offline in a single batch.

 All the folds were executed on a system equipped with an Intel Core i7-13700KF at base clock 3.4 GHz (up to 5.4 GHz in boost), 128 GB of DDR4 RAM, and a NVIDIA RTX4090. 
All the data preprocessing leading to the generation of the LAS examples (for 70 days) took four and a half hours. The data preprocessing (see Figure~\ref{fig:activity}) can scale on multiple threads to extract the data and can offload to the GPU (if present) the heatmap clustering task.

The training used a 5-fold cross-validation approach, by doing so, each fold was composed by 14 days of data (126 examples), leaving enough data for the test fold to cover different seasons.
The overall cross-validation took roughly half an hour not exceeding the 10GB of RAM used.
The LAS task of each fold was noisy, meaning that the framework was not forced to learn a hypothesis covering all examples. As mentioned earlier, uncovered examples contribute a penalty to the final hypothesis score. Specifically, each uncovered example incurs a penalty of 1000. 
The remaining portion of the  hypothesis score, provided through the \texttt{\#bias} directive. As mentioned in Section~\ref{sub:bg}, it is designed according to the \emph{locality principle}, favoring explanations based on humidity, cloud coverage, and temperature at the same location before considering information from neighboring locations. Such a scoring function remains general and can be easily adapted to new regions by defining the notion of neighboring locations accordingly.
Considering this, on the training set our hypotheses covered an average of 98.3\% of the examples across all folds, providing a sufficient accuracy (an average of 85\% over all the folds). Accuracy and recall are reported in Table~\ref{tbl:train}.

\begin{table}[tb]
   \centering
    {\small \begin{tabular*}{0.6\textwidth}{@{\extracolsep{\fill}}cccccc}
    \midrule
         Fold no. & 1 & 2 & 3 & 4 & 5\\
         \midrule
         Accuracy & 0.842 & 0.845 & 0.835 & 0.842 & 0.835 \\
         Recall &  0.848 & 0.853 & 0.823 & 0.856 & 0.813 \\
    \midrule
    \end{tabular*}}
    \caption{The accuracy and recall values obtained on the training set with the 5-fold cross-validation.}
    \label{tbl:train}
\end{table}

\begin{table}[tb]
    \centering
     {\small\begin{tabular*}{0.6\textwidth}{@{\extracolsep{\fill}}cccccccc}
    \midrule
         Fold no. & 1 & 2 & 3 & 4 & 5 \\
         \midrule
         Accuracy & 0.628 & 0.576 & 0.543 & 0.53 & 0.572 \\
         Recall & 0.344 & 0.331 & 0.278 & 0.302 & 0.33 \\
    \midrule
    \end{tabular*}}
    \caption{The accuracy and recall values obtained on the different validation folds.}
    \label{tbl:test}
\end{table}

\paragraph*{Results}
On average, the hypotheses on each validation fold resulted in an accuracy of \textbf{57\%}. While the predictive accuracy may not yet fully satisfactory, the results remain promising given that our main focus is on the explainability provided by the learned rules rather than on performance. In contrast to neural networks, which use high-dimensional representations, logic-based models require an explicit abstraction of the input data, leading to a loss of representational detail. Moreover, due to current scalability limitations, FastLAS can be trained only on relatively small datasets, unlike neural approaches that typically benefit from millions of training examples. Nevertheless, despite these constraints, the learned rules already demonstrate strong explanatory capabilities suggesting that improvements to the model will enhance its accuracy.
The precision and recall values obtained on each validation fold are reported in Table \ref{tbl:test}.
Alongside the metrics obtained by the cross validation, it is relevant to notice that, while present in the search space, none of the five learned hypotheses used facts related to the wind speed (e.g., \lstinline{wind_blowing_morning_speed/2}.). While such facts were not considered  relevant by FastLAS to explain the forecast, in every hypotheses it used  facts concerning the wind direction (especially to derive the predicate \lstinline{forecasted_rain/2}).
As an example consider the following two rules, taken from a \emph{learned} hypothesis:
\begin{lstlisting}
forecasted_sky(trieste,mostly_clear,summer) :- city_covered_less_than(trieste,3), humidity_decreased_at_afternoon(trieste), wind_blowing_morning(trieste,"E").

forecasted_rain(gorizia,1,winter) :- city_covered_less_than(gorizia,3), humidity_decreased_at_afternoon(gorizia), humidity_increased_at_afternoon_neighbour(trieste).
\end{lstlisting}

\noindent
The first it states that in Trieste, during summer, if the prediction says that there are less than~3 hours of  cloud cover and other info on wind and humidity over the city, then we can safely assert, as a summary, that it would be ``mostly clear''. 
The second rule, states that during winter time in Gorizia, a prediction of light rain (rain of level~1) can be justified by a prediction of not many hours of cloud cover (less than~3) and by predicted humidity variations in the towns (locations) of Gorizia and Trieste.

\section{Conclusions}
\label{sec:conclusion}
In this paper we presented a pipeline which, starting from simulated meteorological raw data and from weather bulletins, is able to retrieve ASP facts and generate ILP examples. From such examples we have been able to learn, via FastLAS2, explanatory hypothesis that can be used by readers of the bulletin to better understand some of the reasoning carried out by human experts who produced the actual forecast. Moreover, the hypothesis can also provide high level insights about experts different experience and evaluations of the same scenario. 
Future work will focus on optimizing and generalizing the proposed pipeline. Key refinements will be restructuring the background knowledge base to minimize memory footprint, and leveraging GPU data-parallelism to speed up the hypothesis learning phase \cite{DovierFGHPR22,DovierFV19}. These optimizations will enable the pipeline to process larger/richer datasets. Consequently, system's capabilities will be expanded  by a)~using datasets that include facts for multi-level temperature and humidity data to infer more precise, height-stratified explanations; and b)~using training (and evaluation) datasets that span homologous periods across multiple consecutive years, to account for recurrent multi-year patterns of seasonal phenomena and improve the accuracy of the learned hypothesis.
Additional effort will be put in extending the pipeline to also incorporate additional data sources, and developing it into a fully-fledged framework that can be applied with minimal effort to different geographical regions and types of weather bulletins. We also plan to extend this framework to a complete explainable pipeline, taking into account the final stage that converts concepts (learned rules) to natural language, with a Prolog-based engine as described in~\cite{bertini2024concept2text,bertini2024data2concept2text}. The ultimate goal is to produce automated bulletins that can ease expert's evaluations (see Figure~\ref{fig:sergio}).

\paragraph*{Acknowledgments}
The authors would like to thank all the members of OSMER team, in particular
Sergio Nordio, Agostino Manzato, Federico Bonaldo, and Francesco Sioni,
and the developers of FastLAS  Mark Law,  Matthew Tait, and Alessandra Russo  for their kindness and guidance.

\bibliographystyle{eptcs}
\bibliography{bib}

\end{document}